\documentclass[conference]{IEEEtran}
\IEEEoverridecommandlockouts
\usepackage{cite}
\usepackage{authblk}
\usepackage{amsmath,amssymb,amsfonts}
\usepackage{algorithmic}
\usepackage[ruled,vlined,noend]{algorithm2e}
\usepackage{graphicx}
\usepackage{textcomp}
\usepackage{xcolor}
\usepackage[a4paper, total={184mm,239mm}]{geometry}
\def\BibTeX{{\rm B\kern-.05em{\sc i\kern-.025em b}\kern-.08em
    T\kern-.1667em\lower.7ex\hbox{E}\kern-.125emX}}
\begin{document}

\title{Transformer-based Machine Learning for \\
Fast SAT Solvers and Logic Synthesis
}

% Feng Shi, Chonghan Lee, Mohammad Khairul Bashar, Nikhil Shukla, Song-Chun Zhu and Vijaykrishnan Narayanan 

% \author{
% \IEEEauthorblockN{1\textsuperscript{st} Given Name Surname}
% \and
% \and
% \and
% \and
% \and

% Feng Shi, Chonghan Lee, Mohammad Khairul Bashar, Nikhil Shukla, Song-Chun Zhu and Vijaykrishnan Narayanan 
\author[1]{Feng Shi}
\author[2]{Chonghan Lee} 
\author[3]{Mohammad Khairul Bashar}

\author[3]{\\Nikhil Shukla}
\author[1]{Song-Chun Zhu} 
\author[2]{Vijaykrishnan Narayanan}
\affil[1]{University of California Los Angeles}
\affil[2]{The Pennsylvania State University}
\affil[3]{University of Virginia}

\maketitle

\begin{abstract}
CNF-based SAT and MaxSAT solvers are central to logic synthesis and verification systems. The increasing popularity of these constraint problems in electronic design automation encourages studies on different SAT problems and their properties for further computational efficiency. There has been both theoretical and practical success of modern Conflict-driven clause learning SAT solvers, which allows solving very large industrial instances in a relatively short amount of time. Recently, machine learning approaches provide a new dimension to solving this challenging problem. Neural symbolic models could serve as generic solvers that can be specialized for specific domains based on data without any changes to the structure of the model. In this work, we propose a one-shot model derived from the Transformer architecture to solve the MaxSAT problem, which is the optimization version of SAT where the goal is to satisfy the maximum number of clauses. Our model has a scale-free structure which could process varying size of instances. We use meta-path and self-attention mechanism to capture interactions among homogeneous nodes. We adopt cross-attention mechanisms on the bipartite graph to capture interactions among heterogeneous nodes. We further apply an iterative algorithm to our model to satisfy additional clauses, enabling a solution approaching that of an exact-SAT problem. The attention mechanisms leverage the parallelism for speedup. Our evaluation indicates improved speedup compared to heuristic approaches and improved completion rate compared to machine learning approaches.
\end{abstract}


% --------------------------------- Introduction ----------------------------------
\section{Introduction}

Logic synthesis is a crucial step in design automation systems where abstract logic is transformed to physical gate-level implementation. There has been  significant improvement in hardware performance and cost by optimizing logic at the synthesis level. The task to synthesize and minimize digital circuits is often translated to the Constraint Satisfaction Problem (CSP). CSP aims at finding a consistent assignment of values to variables such that all constraints, which are typically defined over a finite domain, are satisfied. The Boolean Satisfiability (SAT) and Maximum Satisfiability (MaxSAT) solvers have been the core of the Constraint Satisfaction methods to seek a minimal satisfiable representation of logic. Extensive studies have been conducted on MaxSAT problem for logic synthesis \cite{LeiCL20, minimax2007, maxsat08}. 

Previous SAT solvers are based on well-engineered heuristics to search for satisfying assignments. These algorithms focus on solving CSP via backtracking or local search for conflict analysis. David-Putnam-Logemann-Loveland (DPLL) algorithm exploits unit propagation and pure literal elimination to optimize backtracking Conjunctive Normal Form (CNF) \cite{dpll}. Derived from DPLL, conflict-driven clause learning (CDCL) algorithms such as Chaff, GRASP, and MiniSat have been proposed \cite{moskewicz2001chaff, grasp, een03minisat}. Since SAT algorithms can take exponential runtime in the worst case, the search for additional speed up has continued. SAT Sweeping is a method to merge equivalent gates by running simulation and SAT solver in synergy \cite{9218691, 1688794}. MajorSAT proposed efficient SAT solver for solving the instances containing majority functions \cite{7428058}. Another method used directed acyclic graph topology for the Boolean chain to restrict on the search space and reduce runtime \cite{8465888}. The heuristic models improved computational efficiency but are bounded by the greedy strategy, which is sub-optimal in general.

Recently, the machine learning community has seen an increasing interest in applications and optimizations related to neural symbolic, including solving CSP and SAT. With the fast advances in deep neural networks (DNN), various frameworks utilizing diverse methodologies have been proposed, offering new insights into developing CSP/SAT solvers and classifiers. NeuroSAT is a graph neural network model that aims at solving SAT without leveraging the greedy search paradigm \cite{selsam2018learning, selsam2019guiding}. It approaches SAT as a binary classification problem and finds an SAT assignment from the latent representations during inference. The model is able to search for solutions to problems of various difficulties despite training for relatively small number of iterations. As an extension to this line of work, PDP-solver \cite{pdp2019} proposes a deep neural network framework that facilitates satisfiability solution searching within high-performance SAT solvers on real-life problems. However, most of these works, such as neural approaches utilizing RNN or Reinforcement Learning, are still restricted to sequential algorithms, while clauses are parallelizable even though they are strongly correlated through shared variables.

\begin{figure*}[ht]
    \centering
    \includegraphics[width=0.67\textwidth]{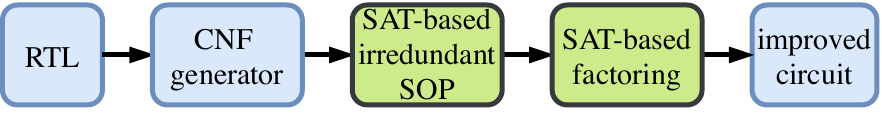}
    \caption{CNF-based SAT solver in logic synthesis flow.}
    \label{fig:ls_flow}
\end{figure*}

In this work, we propose a hybrid model of the \textit{Transformer} architecture \cite{attention_is_all} and the Graph Neural Network for solving CSP/SAT.

Our main contributions in this work are: 
\begin{itemize}
\item We leverage meta-paths, a concept introduced in \cite{sun2011pathsim}, to formulate the message passing mechanism between homogeneous nodes. This enables our model to perform self-attention and pass messages through either variables sharing the same clauses, or clauses that include the same variables. We apply the cross-attention mechanism to perform message exchanges between heterogeneous nodes (i.e., clause to variable, or variable to clause). This enhances the latent features, resulting in better accuracy in terms of the completion rate compared to other state-of-the-art machine learning methods in solving MaxSAT problems.
\item In addition to using a combination of \textit{self-attention} and \textit{cross-attention} mechanism on the bipartite graph structure, we combine the Transformer with Neural Symbolic methods to resolve combinatorial optimization on graphs. Consequently, our model shows a significant speedup in CNF-based logic synthesis compared to heuristic SAT solvers as well as machine learning methods.
\item We propose \textit{Transformer-based SAT Solver} (TRSAT), a general framework for graphs with heterogeneous nodes. In this work, we trained the \textit{TRSAT} framework to approximate the solutions of CSP/SAT. Our model is able to achieve competitive completion rate, parallelism, and generality on CSP/SAT problems with arbitrary sizes. Our approach provides solutions with completion rate of 97\% in general SAT problem and 88\% for circuit problem with significant speed up over prior techniques. 
\end{itemize}

% ---------------------- Background -------------------------------

\begin{table}[ht]
\centering
\resizebox{0.45\textwidth}{!}{
\begin{tabular}{c|p{45mm}}
\hline
Gate  & CNF equation   \\ \hline
$z = a \cdot b$ & $\phi = (a + \neg z) \cdot (b + \neg z) \cdot (\neg a + \neg b + z)$                  \\ %\hline
$z = a + b $ & $\phi = (\neg a + z) \cdot (\neg b + z) \cdot (a + b + \neg z)$ \\ %\hline
$z = \neg a$ & $\phi = (a + z) \cdot (\neg a + \neg z)$        \\ %\hline
$z = a \oplus b$ & $\phi = (a + b + \neg z) \cdot (a + \neg b + z)$ \newline $\qquad\; (\neg a + \neg b + \neg z) \cdot (\neg a + b +  z)$ \\ \hline
\end{tabular}
}
\caption{CNF equations for the basic logic gates \cite{wang2009electronic}}
\label{tab:cnf_equ}
\end{table}

\section{Background} %Preliminaries%
In this section, we introduce the preliminaries for CNF-based logic synthesis and the advanced machine learning models, i.e., Transformers and Graph Neural Networks.  

\subsection{CNF equations for logic gates}

For a logic gate with function $z = f(a, b, \dots)$, it equals to logic expression $(z \Rightarrow f(a, b, \dots)) \cdot (f(a, b, \dots) \Rightarrow z)$, which then derives $(\neg z + f(a, b, \dots)) \cdot (\neg f(a, b, \dots) + z)$. We further expand the above equation in product of sum (POS) form to obtain the CNF for the gate. Table \ref{tab:cnf_equ} summarize the CNF equations for basic logic gates,

\subsection{Transformers and relation to GNNs}
To combine the advantages from both CNNs and RNNs, \cite{attention_is_all} presents a novel architecture, called Transformer, using only the attention mechanism. This architecture achieves parallelization by capturing recurrence sequence with attention and at the same time encodes each item’s position in the sequence. As a result, Transformer leads to a compatible model with significantly shorter training time. The self-attention mechanism of each Transformer layer is depicted as a function $T: \mathbb{R}^{N \times F} \rightarrow \mathbb{R}^{N \times F}$; given $x \in \mathbb{R}^{N \times F}$, the $l$th layer $T_l$ computes,
\begin{align}
    Q = x & W_{Q}, K = x W_{K}, V = x W_{V} \\
    A_l(x) &= V' = softmax(\frac{Q K^T}{\sqrt{D}})V \\
    T_{l}(x) &= f_l(A_l(x) + x) 
\end{align}
where $W_{Q}, W_{K} \in \mathbb{R}^{F \times D}$ and $W_{V} \in \mathbb{R}^{F \times M}$ are projection matrices for evaluating queries $Q$, keys $K$, and values $V$, respectively. $A_l(x)$ are self-attention matrices which describe the similarities between vector entries of $x$. The self-attention matrix $A_l$ is a complete graph which represents the connectivity between queries and keys. When queries and keys are loosely related, the attention map becomes a sparse matrix, similar to the aggregation phase of the Graph Neural Network (GNN). Another difference between the self-attention mechanism used in Transformer and the Graph Attention Network (GAT) \cite{gat2017} is that Transformer's attention mechanism is multiplicative, which is accomplished by dot product, while GAT employs additive attention.

% ------------------------------ Methodology ---------------------------------
% Mat 11:28 Come to Me all who toil and are burdened, and I will give you rest. 
%        29 Take My yoke upon you and learn from Me, for I am meek and lowly in heart, and you will find rest for your souls.
%        30 For My yoke is easy and My burden is light.

\section{Methodology}

\begin{figure}[ht]
    \centering
    \includegraphics[width=0.486\textwidth]{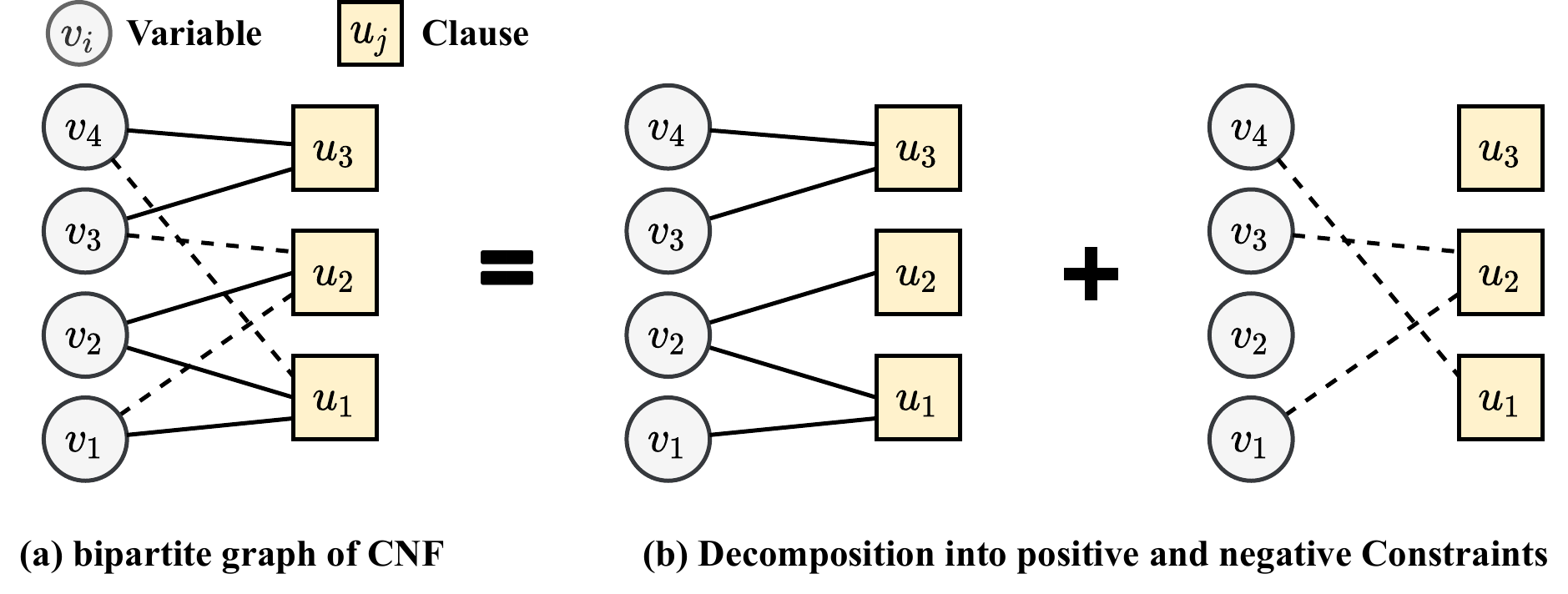}
    \caption{(a) bipartite graph for the \textit{CNF} with measure $\phi = (v_1 \lor v_2 \lor \lnot v_4) \land (\neg v_1 \lor v_2 \lor \neg v_3) \land (v_3 \lor v_4)$, where solid lines are the positive incidences of $v_i$ in $u_j$, and dashed lines are the negative incidences of $\neg v_i$ in $u_j$; (b) the decomposition of the bipartite graph according to the positive and negative relations.}
    \label{fig:cnf_factor}
\end{figure}

\begin{figure*}[ht]
    \centering
    \includegraphics[width=0.895\textwidth]{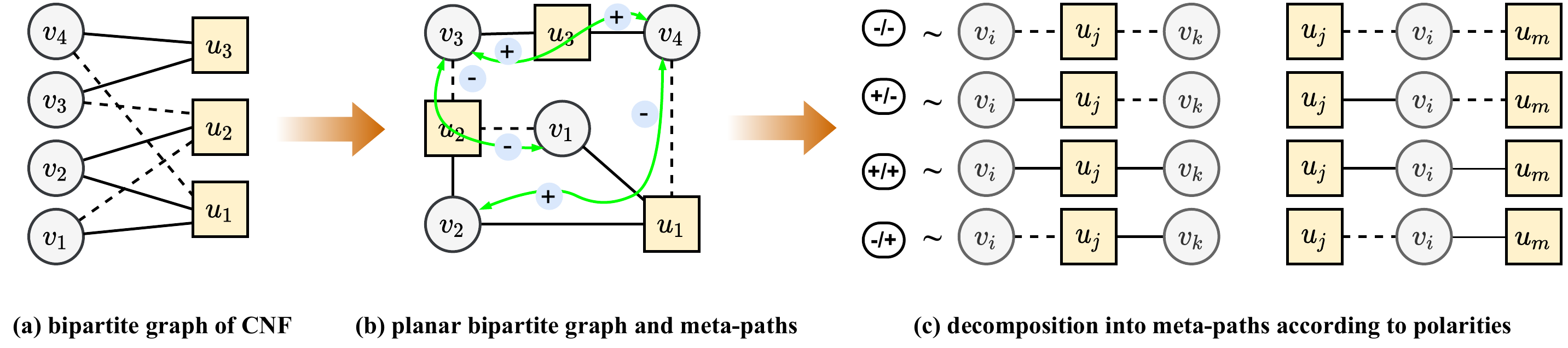}
    \caption{left: bipartite graph from Fig.\ref{fig:cnf_factor}(a); right: planar topology of bipartite graph and the meta-paths marked with $\{+, -\}$}
    \label{fig:meta_paths}
\end{figure*}

In this section, we present the methodologies applied in this work. Specially, we discuss the graph representation of CNF in Section \ref{sec:bipartite}, a flexible sparse attention for both self- and cross-attention in Section \ref{sec:sparse_attn}, and the overall architecture of framework in Section \ref{sec:overall}.

\begin{figure}[ht]
    \centering
    \includegraphics[width=0.46\textwidth]{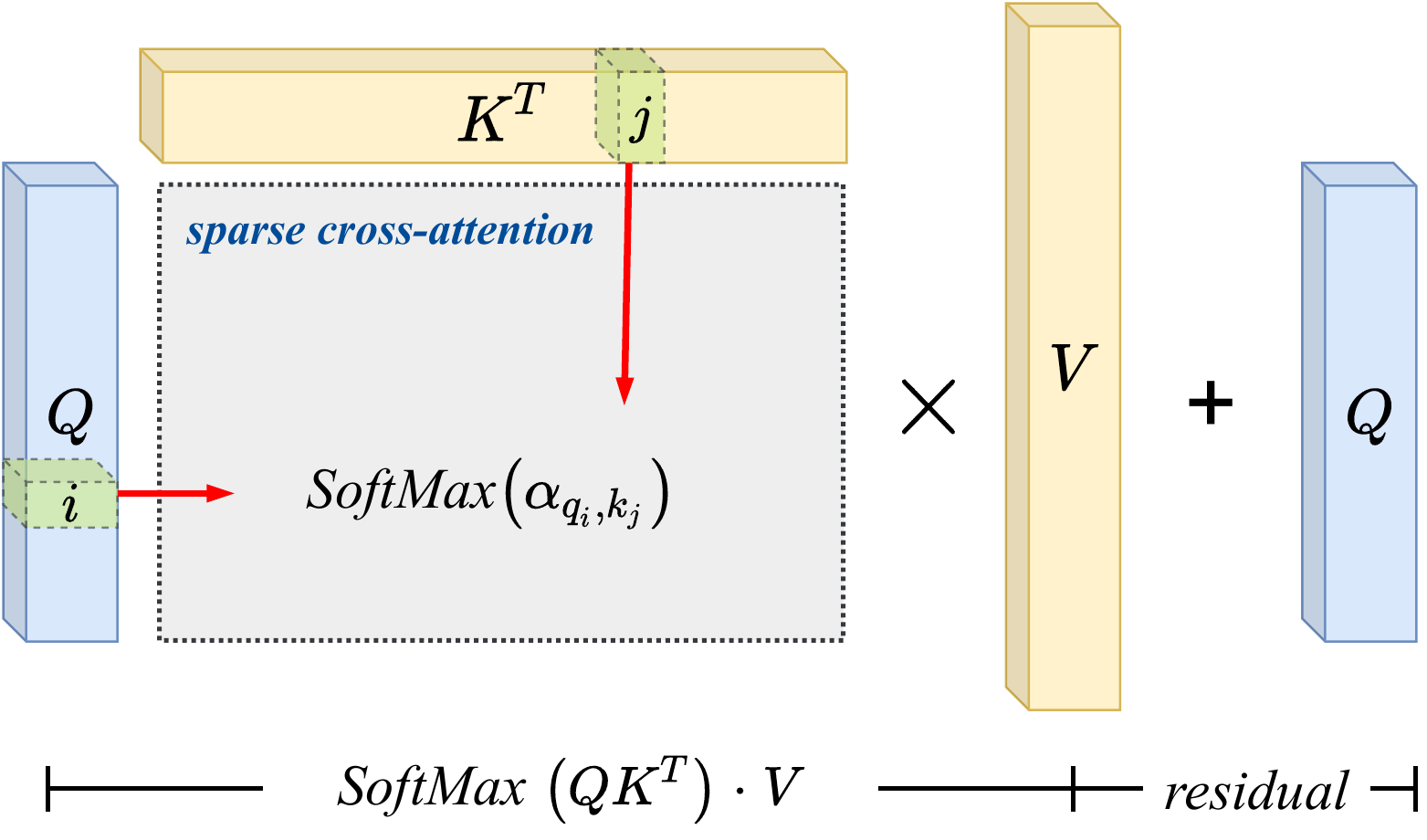}
    \caption{Sparse attention mechanism}
    \label{fig:cross_attn}
\end{figure}
\vspace{-3pt}
\subsection{CNF as bipartite graph and the concept of meta-paths} \label{sec:bipartite}

Each \textit{CNF} equation can be formulated as,

\begin{equation}
    \phi(V, U) = \prod_{j=1}^{M}\sum_{i \in u_{j}} \{v_i \text{ or } \neg v_i \}\text{ for } v_i \in V \text{ and } u_j \in U
\end{equation}

where $V$ and $U$ are the sets of variables and clauses, respectively. Either variable $v_i$ or $\neg v_i$ appears in the clause $u_j$, but not both at the same time. The expression can be properly presented as an undirected bipartite graph, as shown in Fig.\ref{fig:cnf_factor}(a). We then construct such a bipartite graph $G((V, U), \mathcal{E})$ by defining the set of variables $V=\{v_1,\dots, v_n\}$, the set of clauses $U=\{u_1,\dots, u_m \}$, and edges $\mathcal{E}$ by: $e_{i,j} \in \mathcal{E}$ iff variable $v_i$ is involved in constraint $u_j$ either in positive or negative relation. To assist the message passing mechanism used in graph neural network, we further separate the bipartite graph in two sub-graphs, one for positive constraints and another for negative constraints, e.g., $\phi_{+} = (v_1 \lor v_2)_{u_1} \cdot (v_2)_{u_2} \cdot (v_3 \lor v_4)$ and $\phi_{-} = (\neg v_2 \lor \neg v_4) \cdot (\neg v_1)$. Moreover, given the adjacency matrix $A$ of the bipartite graph, each edge is assigned with a type depending on the polarity of the variable to which it connects. The positive occurrence of a variable $v_i$ in a clause (or factor) $u_j$ is represented with the positive sign $(+)$, whereas its negative occurrence $\neg v_i$ in $u_{j}$ gets the negative sign $(-)$. Hence, a pair of $n \times m$ bi-adjacency matrix $\textbf{A} = (A_{+}, A_{-})$, which correspond to the pair $(\phi_{+}, \phi_{-})$, is used to store two types of edges such that $A_{+}(i,j)=1 \Leftrightarrow {v_i \in u_j}$ and $A_{-}(i,j) = 1 \Leftrightarrow {\neg v_i \in u_j}$, the example of decomposed sub-graphs are shown in Fig.\ref{fig:cnf_factor}(b). Here $v_i \in u_j$ implies that $v_i$ instead of its negation $\neg v_i$ is directly involved in $u_j$. Each edge $e_{i,j} \in \mathcal{E}$ is then assigned a value equal to $1$ for edges in $A_{+}$ and $-1$ for edges in $A_{-}$. With the graph representation, graph neural network can be applied to solve symbolic reasoning problem, e.g., CSP/SAT-solver \cite{selsam2018learning, rlsat19}. These two sub-graphs are then applied with the self-attention on positive and negative links, i.e., the positive and negative constraints in CNFs, respectively, as explained in Section \ref{sec:sparse_attn}.

Due to bipartite properties, variables are only connected to clauses, and vice versa, as shown in Figure \ref{fig:cnf_factor}. 
Consequently, every node must traverse a node with different type to reach a node with same type. Furthermore, traditional \textit{GNN} can only transfer messages between nodes with the same attributes. In this work, we propose to pass message through 2-hop \textbf{meta-paths} \cite{sun2011pathsim} in addition to existing edges, which enables variables (clauses) to incorporate the information from variables (clauses) that share the same clauses (variables) during the update of their states. In a \textit{CSP/SAT} factor graph, we define that a \textit{meta-path} $m_{i,j} = (v_i, u_k, v_j)$ between nodes $v_i$ and $v_j$ exists if there exists some $u_k \in U$ s.t. $\exists e_{i,k} \in \mathcal{E}$ and $\exists e_{k,j} \in \mathcal{E}$. Since self-attention mechanism is not symmetric, our meta-path is directed. As a result, we get four types of meta-paths in total, i.e., $\{(+, +), (+, -), (-, +), (-, -)\}$, as illustrated in right-hand side of Fig.\ref{fig:meta_paths}. The adjacent matrix of such a meta-path can be easily computed by matrix multiplication of $A_+$ and $A_{-}$ or their transposes. Take $A_{(+,+)}$, $A_{(+,-)}$ as examples, the adjacency matrix $A_{(+,+)} = A_{+} A_{+}^T$ stores all $(+, +)$ meta-paths, and $A_{(+,-)} = A_{+} A_{-}^T$ stores all $(+,-)$ meta-paths. A diagonal entry $A_{(+, +)}[i, i]$ indicates the number of positive edges that $v_i$ has, and an off-diagonal entry $A_{(+,+)}[i, j]$ indicates the existence of $(+, +)$ meta-path from $v_i$ to $v_j$.

\begin{figure*}[ht]
    \centering
    \includegraphics[width=0.897\textwidth]{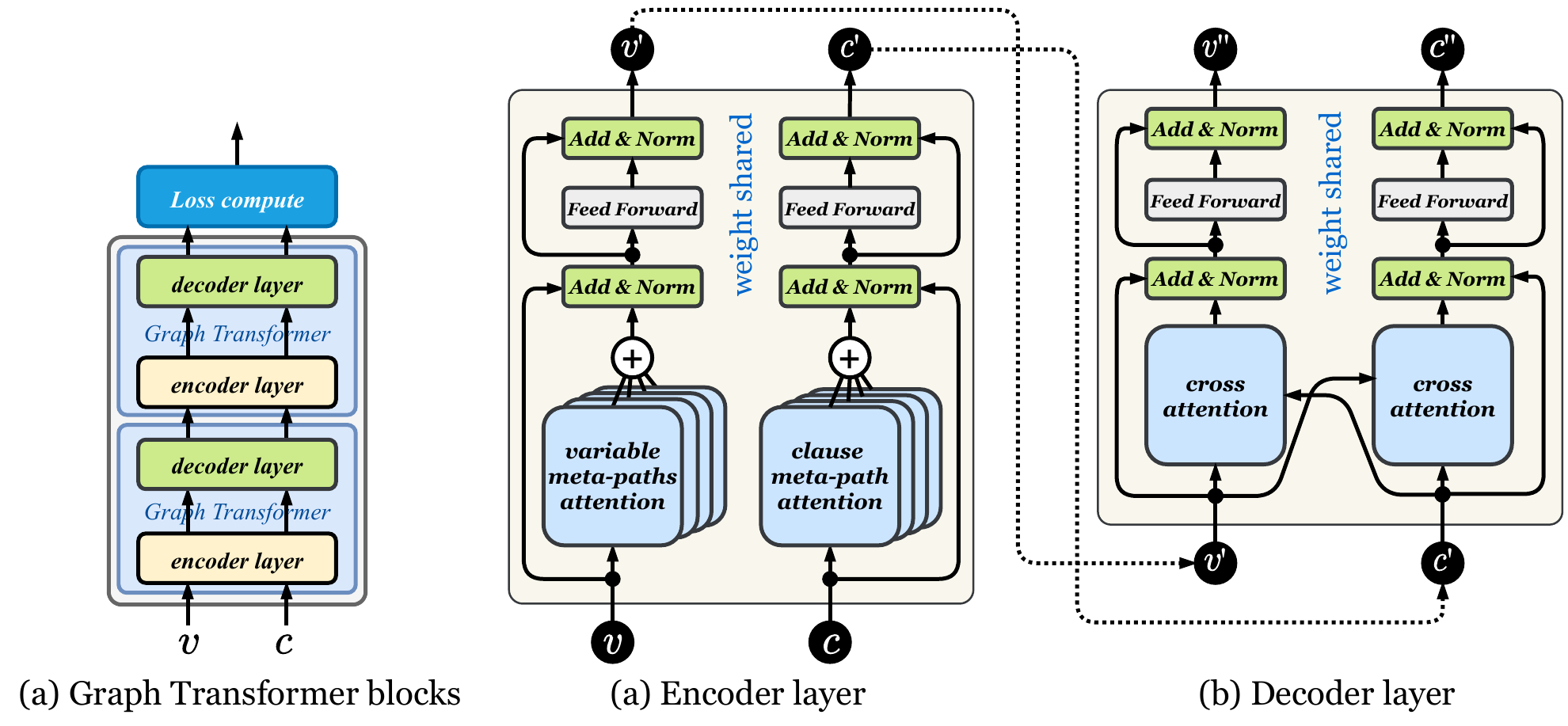}
    \caption{Our \textit{Transformer-based SAT} (TRSAT) solver architecture consists of a set of encoders and decoders connected sequentially, as in (a). Its encoder and decoder architectures are shown in (b) and (c), respectively.}
    \label{fig:gtr_arch}
\end{figure*}

\subsection{Sparse attention and graph Transformer} \label{sec:sparse_attn}

This work employs sparse attention coefficients for both the self-attention of meta-paths and the cross-attention between variables and clauses, as explained in section \ref{sec:overall}. The sparse attention coefficient is calculated according to the connectivity between graph nodes. As described in below equations: after the embedding in Equation \ref{eq:embedding}, where $X = Y$ for self-attention and $X \neq Y$ for cross-attention, as shown in Figure \ref{fig:cross_attn}. The node-to-node attention between a pair of connected nodes is computed first by an exponential score of the dot product of the feature vectors of these two nodes (Equation \ref{eq:attn}). Then the score is normalized by the sum of exponential scores of all neighboring nodes as described in Equation \ref{eq:sparse_attn}.
\begin{align}
    Q &= X W_q, K = Y W_k, V = Y W_v \label{eq:embedding}\\
    \alpha_{i, j} &= \frac{\left<q_i, k_j\right>}{\sum_{n \in N(i)} \left< q_i, k_n \right>} \text{ or } SM(\left<Q[rows], K[cols] \right>) 
   \label{eq:attn} \\
    v_{i}' &= \sum_{j \in N(i)} \alpha_{i, j} v_j, \quad\text{ or }  V' = \underbrace{\mathcal{A} }_{sparse} \times V
    \label{eq:sparse_attn}
\end{align}

where $Q$, $K$, and $V$ are queries, keys, and values in Transformer's terminology, respectively; and $q_i = Q(i)$, $k_j = K(j)$, $v_j = V(j)$, $\left< q, k \right> = exp\left( \frac{q^T k}{\sqrt{d}} \right)$, and $SM(\cdot)$ is the \textit{SoftMax} operation. Finally, we obtain attention maps $\mathcal{A}$ as multi-dimension (multi-head) sparse matrices sharing the identical topology described by a single adjacency matrix, where attention coefficients are $\mathcal{A}(i, j) = \alpha_{i, j}$. The sparse matrix multiplications can be efficiently implemented in high parallelism with the tensorization of node feature gathering and scattering operations through indexation.

\subsection{Loss Evaluation}
For a given $\text{SAT}(V, U)$, each combination of variable assignments corresponds to a probability. The original measure $\phi (V, U)$ is a non-differentiable staircase function defined on a discrete domain. $\phi (V, U)$ evaluates to $0$ if any $u_j \in U$ is unsatisfied, which disguises all other information including the number of satisfied clauses. For training purpose, a differentiable approximate function is desirable. Therefore, the proposed model generates a continuous scalar output $x_i \in [0, 1]$ for each variable, and the assignment of each $v_i$ can be acquired through:
\begin{equation}
    v_i = \Bigl\lfloor\dfrac{x_i}{0.5 + \epsilon}\Bigr\rfloor
\end{equation}
where $\epsilon$ is a small value to keep the generated $v_i$ in $\{0, 1\}$. With continuous $x_i, i = {1,...,N}$, we can approximate disjunction with $max(\cdot)$ function and define $\phi(\cdot)$ as
\begin{equation}
    \phi(x_1,...,x_N) = \prod_{j=1}^M \max (\{l(x_i) : v_i \in u_j\})
\end{equation}
Here, the literal function $l(x_i,e_{ia})=\frac{1-e_{ia}}{2} + e_{ia}x_i$ is applied to specify the polarity of each variable. % Inspired by \cite{pdp2019}, we 
We replace the \textit{max} function with a differentiable \textit{smoothmax}, $S_r(\cdot)$:
\begin{equation}
    S_{\tau}(x_1,...,x_N) = \frac{\sum_{i=1}^n x_ie^{\tau x_i}}{\sum_{i=1}^n e^{\tau x_i}}
\end{equation}
Mathematically, $S_{\tau}(x_1,...,x_N)$ converges to $\max(x_1,...,x_N)$ as $\tau \rightarrow \infty$. 
We note that $\tau = 5$ is enough for our model in practice. By maximizing the modified $\phi$, the proposed model is trained to find the satisfiable assignment for each CSP problem. For numerical stability and computational efficiency, we train our model by minimizing the \textit{negative log-loss}

\begin{equation}
    \mathcal{L}(x_i,...,x_N) = %-log(\phi) =
    -\sum_{j=1}^{M}log(S_{\tau}(\{l(x_i) : v_i \in u_j\}))
\end{equation}

\subsection{Heterogeneous Graph Transformer Architecture} \label{sec:overall}

We further propose the Heterogeneous Graph Transformer (TRSAT) which adopts an encoder-decoder structure, as illustrated in Figure \ref{fig:gtr_arch}(a). It is a flexible architecture allowing the number of encoder- and decoder-layers to be adjustable.

\textbf{Encoder}. Within each encoder-layer, every graph node first aggregates the message (or information) from nodes of its kind through \textit{meta-paths}. Note that a node (variable or clause) of bipartite graph has no direct connection within homogeneous nodes. Messages can only pass among homogeneous nodes through \textit{meta-paths}. We emphasize such type of communication between nodes of the same kind as \textit{self-attention}, which is implemented with \textit{homogeneous attention} mechanism regarding the polarity of variables. The attention are then connected to the residual block and \textit{layer normalization} \cite{layernorm_ba2016}, as shown in Figure \ref{fig:gtr_arch} (b).

\textbf{Decoder}. Inside each decoder-layer, the weighted messages are passed between variables and clauses through the \textit{cross-attention} mechanism, implemented as the \textit{heterogeneous attention} regarding nature of graph nodes (either variables or clauses), followed by residual connection and \textit{layer normalization}, as in Figure \ref{fig:gtr_arch} (c). The attention-weighted node features are then fed into the feed-forward network (\textit{FFN}) for enhancing the node feature embedding.

\subsection{Analysis of the Complexity}

We initiate the discussion of the complexity from computing single attention head, multi-head follows the same analysis. Both the self-attention of meta-paths and the cross-attention between variables and clauses described in previous sections rely on the connectivity (or topology) of the relevant bipartite graphs, so the time complexity of computing these attention coefficients is $\mathcal{O}(|\mathcal{E}| \times |F|)$, where $|\mathcal{E}|$ is the number of edges in a graph and $|F|$ is the number of features of graph node. The node encoding module, which is a linear layer in the model, and the feed-forward network (FFN) module possess the time complexity of $\mathcal{O}(|V| \times F \times F')$, for $|V|$ the number of graph nodes. As $|\mathcal{E}| \gg |F|$ and $|V| \gg |F|$, total complexity of a single attention head is proportional to the number of nodes and edges. Furthermore, space complexity of the memory footprint for sparse attention is also linear in terms of nodes and edges.

\subsection{MaxSAT approximates Exact SAT}
Depends on the application's requirement, the SAT problem can be further categorized as the \textit{maximum satisfiability problem (MAX-SAT)} and \textit{exact SAT}. MaxSAT determines the maximum number of clauses of a given Boolean formula in Conjunctive Normal Form (CNF), which can be made true by an assignment of truth values to the formula's variables  \cite{enwiki-maxsat}. It is a generalization of the Boolean satisfiability problem (exact SAT), asking whether a truth assignment makes all clauses valid. Machine learning-based algorithms explore the solution space by minimizing the loss to ground truth and updating their models' weights through gradient descent during the training phase. This constraint has naturally drawn the machine learning-based approaches to focus on MaxSAT problems by performing probabilistic decision-making. Rather than obtaining the deterministic and complete solution, they approximate variable assignments. To remedy this drawback, iterative algorithms can be applied. Different from the \textbf{decimation} strategy employed in PDP, which selectively fixes the values of variables of the solved clauses, we deliver \textbf{Algorithm \ref{algo:max_2_exact}} which conditionally removes the solved clauses and their related variables from current problem. As the decimation approach of PDP does not reduce the problem's scale by fixing values of variables, our model can generate a faster and more efficient solution by decreasing the size of the problem.

\begin{algorithm}[h]
\SetAlgoLined
\DontPrintSemicolon % Some LaTeX compilers require you to use \dontprintsemicolon instead
\KwIn{
    $V = \{ v_1, v_2, \cdots, v_N \}$, 
    $C = \{ c_1, c_2, \cdots, c_M \}$, and 
    $\textit{VAR}(c_i)$: set of all variable in the clause $c_i$
}
\KwOut{
    $V_{sat}: \{v_i = 0 \text{ or } 1\}^N_{i=1}: \text{ assignments of V}$
}
$V_{sat} \gets \text{\O}; U \gets C$; \\
\For{$U \neq \text{\O}$}{
    $(V, C) = \textit{TRSAT}(V, C)$ \tcp*{our model performs one-shot MaxSAT}
    $U \gets \{c_i = 0: c_i \in C\}$ \tcp*{get unsolved clauses}
    \eIf{$U = \text{\O}$}{
        \Return{$V_{sat} \cup \{v_i \in V\}$} 
    }{
        $V_u \gets \cup  \{\textit{VAR}(c_i): c_i \in U\}$ \\
        \eIf{$V - V_u = \text{\O}$}{
            \Return{$solvable \gets \textbf{false}$}
        }( \tcc*[h]{conditionally remove the satisfied clauses} ){
            $ C \gets C - \{ c_i = 1: \exists v_j \in c_i \text{ and } v_j \notin V_u \} $ \\
            $ V \gets V_u \cup \{ \textit{VAR}(c_i): c_i \in C \}$ \\
            $V_{sat} \gets V_{sat} \cup (V - V_u)$
        }
    }
}
\Return{$V_{sat}$}
\caption{Iterative algorithm MaxSAT approximates exact-SAT}
\label{algo:max_2_exact}
\end{algorithm}

% ------------------------------ Experiments ---------------------------------
% \input{_tab_k_rand_summary}
\begin{table*}[h!]
\centering
%  \caption{ The summary of our chosen dataset. For random k-SAT problems, \textit{n} and \textit{m} refer to the number of variables and clauses.  For graph problems, \textit{N} is the number of vertices, \textit{k} is the problem-specific parameter, and \textit{p} is the probability that an edge exists.}
% \resizebox{\columnwidth}{!}{
\resizebox{0.65\textwidth}{!}{
\begin{tabular}{cccc}
\hline
\textbf{Class} & \textbf{Distribution} & \textbf{Variables (n)} & \textbf{Clauses} (m) / \textbf{Edges} ($p$) \\ \hline
\textbf{Random 3-SAT} & $rand_3(n,m)$ & $n = \{100, 150, 200\}$ & $m = \{430, 645, 860\}$ \\ \hline
\textbf{$k$-coloring} & $color_{k} (N,p)$ & $n = k \times N$ for $N$=$\{5, 10\}$ & $p = 50\%$ \\ \hline
\textbf{$k$-cover} & $cover_{k}(N,p)$ & $n$=$(k$ + $1) \times N$ for $N$=$\{5, 7\}$ & $p$ = $50\%$ \\ \hline
\textbf{$k$-clique} & $clique_{k}(N,p)$ & $n = k \times N$ for $N$=$\{5, 10\}$ & $p = \{20\%, 10\%\}$ \\ \hline

\end{tabular} 
}

 \caption{ The summary of our chosen dataset. For random k-SAT problems, \textit{n} and \textit{m} refer to the number of variables and clauses.  For graph problems, \textit{N} is the number of vertices, \textit{k} is the problem-specific parameter, and \textit{p} is the probability that an edge exists.}
 
\label{tab:k_rand_dataset}
\end{table*}

% \begin{table*}[h!]
% \centering
%  \caption{ The summary of our chosen dataset. For random k-SAT problems, \textit{n} and \textit{m} refer to the number of variables and clauses.  For graph problems, \textit{N} is the number of vertices, \textit{k} is the problem-specific parameter, and \textit{p} is the probability that an edge exists.}
% \begin{tabular}{cccc}
% \hline
% \textbf{Class} & \textbf{Distribution} & \textbf{Variables (n)} & \textbf{Clauses} (m) / \textbf{Edges} ($p$) \\ \hline
% \textbf{Random 3-SAT} & $rand_3(n,m)$ & $n = \{100, 150, 200\}$ & $m = \{430, 645, 860\}$ \\ \hline
% \textbf{$k$-coloring} & $color_{k} (N,p)$ & $n = k \times N$ for $N$=$\{5, 10\}$ & $p = 50\%$ \\ \hline
% \textbf{$k$-cover} & $cover_{k}(N,p)$ & $n$=$(k$ + $1) \times N$ for $N$=$\{5, 7\}$ & $p$ = $50\%$ \\ \hline
% \textbf{$k$-clique} & $clique_{k}(N,p)$ & $n = k \times N$ for $N$=$\{5, 10\}$ & $p = \{20\%, 10\%\}$ \\ \hline
% \end{tabular}
% \label{tab:circuit_dataset}
% \end{table*}

\section{Experiments Evaluation}

\subsection{Dataset}
To learn a CSP/SAT solver that can be applied to diverse classes of satisfiability problems, we selected our training set from four classes of problems with distinct distributions: random 3-SAT, graph coloring (k-coloring), vertex cover (k-cover), and clique detection (k-clique). For the random 3-SAT problems, we used 1200 synthetic SAT formulas in total from the \textit{SATLIB} benchmark library \cite{hoos2000satlib}. These graphs, consisting of variables and clauses of various sizes, should reflect a wide range of difficulties. For the latter three graph-specific problems, we sampled 4000 instances from each of the distributions that are generated according to the scheme proposed in \cite{rlsat19}.

For evaluating our model's performance on CNF-based logic synthesis, we collected several circuit datasets \cite{circuit_sat_lib, junttila2000towards}, including various Data Encryption Standard (DES) circuits and arithmetic circuits, from real-life hardware designs, and translated them into their corresponding \textit{CNF} formats. Each dataset consists of 100 to 200 samples. Each benchmark subfamily of the DES circuit models, denoted as $des\_r\_b$, is parameterized by the number of rounds ($r$) and the number of plain-text blocks ($b$). The selected arithmetic circuits consist classical adder-tree (atree) as well as Braun multipliers (braun), and are denoted by their names. The largest instances from these circuit dataset contain 14K variables and 42K clauses on average, which is comparable to medium-sized SAT competition instances \cite{sat2018com}.

\subsection{Baselines}
\textbf{Baseline models}. To fully assess the validity and performance of our model in both CSP/SAT solving and CNF-based logic synthesis, we compared our framework against three main categories of baselines: (a) the classic stochastic local search algorithms for SAT solving - \textbf{WalkSAT} \cite{walksat} and \textbf{Glucose} \cite{glucose09} (a variant of \textbf{MiniSAT} \cite{een03minisat}), (b) the reinforcement learning-based SAT solver with graph neural network used for embedding phase - \textbf{RLSAT} \cite{rlsat19}, (c) the generic but innovative graph neural framework for learning CSP solvers -  \textbf{PDP} \cite{pdp2019}. Among these baselines, \textbf{PDP} falls into the hybrid of recurrent neural network and graph neural network based one-shot algorithm. 

\begin{table*}[h]
\centering
\resizebox{0.67\textwidth}{!}{
\begin{tabular*}{10.3cm}{ccccccc}
\cline{1-7}
\multicolumn{1}{l}{\textbf{}} &
  \begin{tabular}[c]{@{}c@{}}$color_3$ \\ $(5, 0.5)$ \end{tabular} & % $color_3(5, 0.5)$ &
  \begin{tabular}[c]{@{}c@{}}$color_3$ \\ $(10, 0.5)$ \end{tabular} & % $color_3(10, 0.5)$ &
  \begin{tabular}[c]{@{}c@{}}$cover_2$ \\ $(5, 0.5)$ \end{tabular} & %$cover_2(5, 0.5)$ &
  \begin{tabular}[c]{@{}c@{}}$cover_3$ \\ $(7, 0.5)$ \end{tabular} & %$cover_3(7, 0.5)$ &
  \begin{tabular}[c]{@{}c@{}}$clique_3$ \\ $(5, 0.2)$ \end{tabular} & %$clique_3(5, 0.2)$ &
  \begin{tabular}[c]{@{}c@{}}$clique_3$ \\ $(10, 0.1)$ \end{tabular} \\  %$clique_3(10, 0.1)$ \\
  \hline
\multicolumn{1}{c}{RLSAT} &
  \begin{tabular}[c]{@{}c@{}}99.01\%\\ $\pm$9.93\%\end{tabular} &
  \begin{tabular}[c]{@{}c@{}}71.49\%\\ $\pm$20.92\%\end{tabular} &
  \begin{tabular}[c]{@{}c@{}}93.78\%\\ $\pm$12.12\%\end{tabular} &
  \begin{tabular}[c]{@{}c@{}}92.04\%\\ $\pm$13.76\%\end{tabular} &
  \begin{tabular}[c]{@{}c@{}}95.72\%\\ $\pm$15.40\%\end{tabular} &
  \begin{tabular}[c]{@{}c@{}}97.96\%\\ $\pm$12.35\%\end{tabular} \\ 
  \hline
\multicolumn{1}{c}{Ours} &
  \multicolumn{2}{c}{\textbf{87.51\%$\pm$1.45\%}} &
  \multicolumn{2}{c}{\textbf{97.77\%$\pm$0.11\%}} &
  \multicolumn{2}{c}{\textbf{97.35\%$\pm$0.37\%}} \\ \hline
\end{tabular*} 
}
\caption{Our model TRSAT's solver performance compared to that of the baseline model RLSAT. We present the metric of percentage completion in the format of: [avg.completion rate]$\pm$[std. deviation]\%.
}
\label{tab:result_3k}
\end{table*}

\subsection{Experimental Configuration}

\textbf{Hardware}. Every experiment is performed on the system with AMD Ryzen 7 3700X 8 core 16 threads CPU equipped with GeForce RTX 2080 8 GB of memory GPU. Since RLSAT, PDP, and our model consists of paralizable operations, we fully deployed on the GPU. Glucose and WalkSAT, which are sequential algorithms using backtracking, are unable to exploit the GPU. 
%do not leverage the benefit of the GPU 

\textbf{Software}. Our model is implemented with PyTorch deep learning framework and employs PyTorch Geometric \cite{pyg2019} for graph representation learning, and is able to achieve high efficiency in both training and testing by taking full advantage of GPU computation resources via parallelism. 

\begin{table*}[h]
\centering
% \caption{ Our model GT's solver performance compared to that of the baseline model PDP \cite{pdp2019}. We present the validation accuracy in the format of: [avg. accuracy \%]$\pm$[std. deviation \%].
%     % [average accuracy\%]$\pm$[standard deviation\%]. 
%     }
% \resizebox{\columnwidth}{!}{
\resizebox{0.7\textwidth}{!}{
\begin{tabular}{ccccccc}
\hline
          & \multicolumn{2}{c}{$rand_3(100, 430)$}              & \multicolumn{2}{c}{$rand_3(150, 645)$}              & \multicolumn{2}{c}{$rand_3(200, 860)$} \\ \hline
          & Time (s)         & Acc (\%)             & Time (s)         & Acc (\%)             & Time (s)    & Acc (\%)     \\ \hline
PDP       & 0.0743           & 96.51$\pm$0.69          & 0.0413           & 95.50$\pm$0.23          & 0.0915      & 93.65$\pm$0.62  \\ \hline
Ours & \textbf{0.00368} & \textbf{97.06$\pm$0.28} & \textbf{0.00361} & \textbf{96.80$\pm$1.31} & \textbf{0.0128}   & \textbf{96.19$\pm$1.57}    \\ \hline
\end{tabular} 
}
    \caption{ Our model TRSAT's performance compared to that of the baseline model PDP \cite{pdp2019}. We present the validation accuracy (completion rate) in the format of: [avg.accuracy]$\pm$[std. deviation] \%.
    % [average accuracy\%]$\pm$[standard deviation\%]. 
    }
    \label{tab:result_uf}
\end{table*}

\textbf{General setup}. Our model for the experiments discussed in this section is configured as follows. Structures are implemented according to the architecture presented in Figure \ref{fig:gtr_arch}. For the encoder, we adopted four layers for both the \textit{encoder-layers} and \textit{decoder-layers} with the number of channels setting to 64 for all of them. Optimizer Adam \cite{kingma2017adam} with $\beta_1 = 0.9$, $\beta_2 = 0.98$, $\epsilon = 10^{-9}$ was applied to train the model. Our learning rate varies with each step taken, and follows a pattern that is similar to the one adopted by Noam \cite{attention_is_all}.
% \begin{equation}
%     learning\_rate= \beta * \min (step\_num^{-0.5}, step\_num \times warmup\_steps^{-1.5})
% \end{equation}
% Specifically, our learning rate increases linearly for the initial warm-up steps, then gradually decays at a rate that is inversely proportional to the square root of the step number. 
% Compared to the PDP solver, our model employs less trainable parameters. Therefore, we enforce a weight decay of 0.01, which is significantly larger than that used in a PDP solver. To ensure fairness of comparison, we adopted a dropout rate of 0.2 for regularization, which is the same value as \cite{pdp2019}.

% \input{_tab_sat_circuit}

% \begin{table*}[!h]
% \centering
% \resizebox{0.583\textwidth}{!}{
% \begin{tabular}{cccc}
% \hline
% \textbf{Model}    & \textbf{CUDA time (ms)} & \textbf{num. of parameters} & \textbf{GMACs}\\ \hline
% RLSAT   & 333.89  & 3.1M               & 261.49\\
% PDP    & 1593.05  & 6.73M              & 197.55\\
% TRSAT  & 86.54   & 0.51M              & 15.58\\ \hline
% \end{tabular}
% }
% \caption{Comparison of models' efficiency}
% \label{tab:efficiency}
% \end{table*}

\begin{table*}[h]
\centering
\resizebox{0.657\textwidth}{!}{

\begin{tabular*}{10cm}{ccc|ccc|c|c}
\multicolumn{3}{c|}{\textbf{SAT Dataset}} & \multicolumn{3}{c|}{\textbf{Ours}} & \textbf{Glucose} & \textbf{WalkSAT} \\ \hline
Data & $\overline{\#V}$ & $\overline{\#C}$ & $\overline{p\%}$ & $\overline{t}(CPU)$ & $\overline{t} (GPU)$ & $\overline{t}$ & $\overline{t} $ \\ \hline
des-3-1 & 5181  & 15455 & 90.7 & 1.56 & 0.113  & 0.73   & 1.26 \\
des-4-1 & 7984  & 23944 & 89.5 & 6.77 & 1.102  & 6.36  & 6.14 \\
des-4-2 & 14027 & 42232 & 87.4 & 7.45 & 1.396  & 6.17  & 7.33 \\ \hline
atree   & 13031 & 41335 & 88.8 & 8.61 & 2.396  & 7.14 & 8.50 \\
braun   & 4116  & 13311 & 92.7 & 6.37 & 1.166  & 5.06 & 6.68 \\   
\hline        
\end{tabular*}}
\caption{Average test completion rate ($\overline{p}\% = \frac{solved\, samples}{\# SAT\, samples}$) and average solving time ($\overline{t}$ seconds) between our model and other approaches for CNF-based logic synthesis.
}
\label{tab:circuit_sat}
\end{table*}

\subsection{Results and Evaluation}
\begin{figure}[h]
    \centering
    \includegraphics[width=0.49\textwidth]{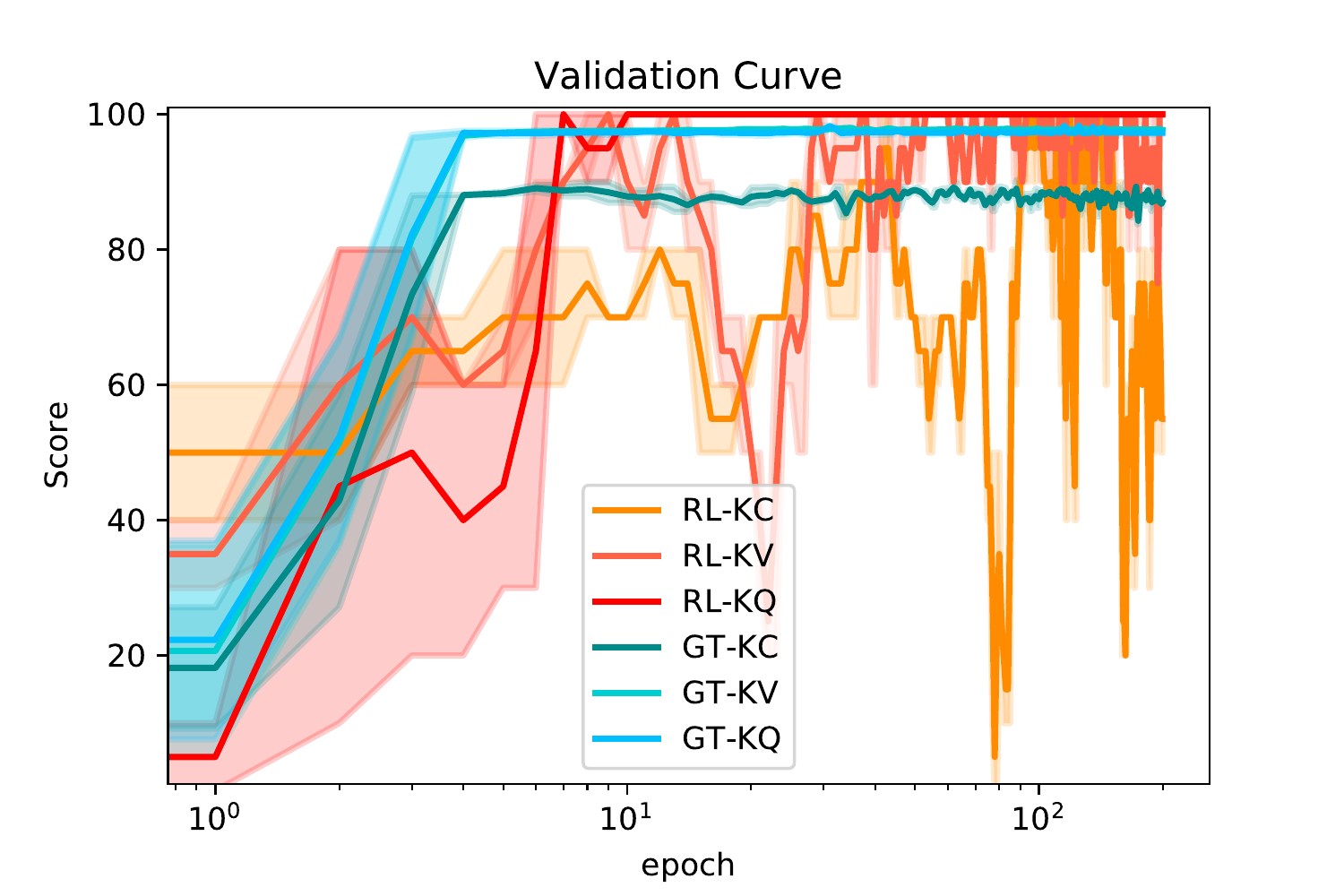}
        \caption{The learning curve of Our model (TRSAT) and that of RLSAT (RL).}
        \label{fig:validation_curve}
\end{figure}

\begin{figure}[h]
    \centering
    \includegraphics[width=0.48\textwidth]{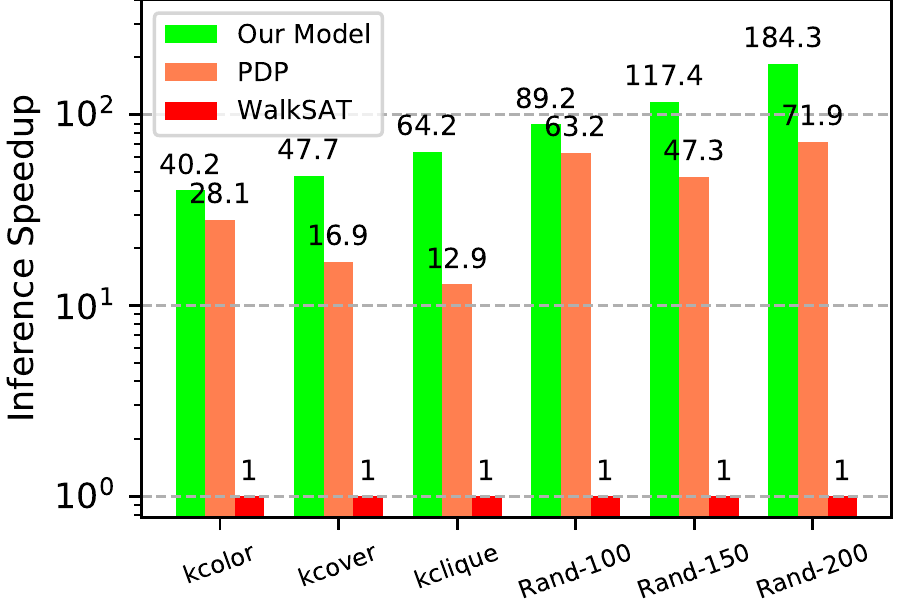} 
    \caption{
    The inference speedup for CSP/SAT solving. }
    \label{fig:bar_chart}
\end{figure}

\subsubsection{\textbf{General CSP/SAT solving}} 
% To demonstrate our model's validity as a SAT solver, we present the following results to demonstrate our performance on general CSP/SAT solving.
We first compare the accuracy metric with RL-based deep model. The accuracy metric represents the average percentage of clauses solved by the models with the generated assignments to variables. Due to the sequential nature of RL, the runtime performance compared to our model is not insightful. Table \ref{tab:result_3k} summarizes the performance of our model and that of RLSAT, after training for 500 epochs, on the chosen datasets. Since our model adopts a semi-supervised training strategy, and is capable of processing graphs of arbitrary size, we were able to combine numerous distributions of the same problem class into one single dataset during training, regardless of the problem-specific parameters. Our model achieves higher completion rate than RLSAT. 
% We observed that our model's average accuracy is higher than that of RLSAT in all six distributions. 

For further analysis, we present the holistic learning curves in Figure \ref{fig:validation_curve}. In this figure, both models are trained on \{KC: $color_3(10, 0.5)$, KV: $cover_3(7, 0.5)$, KQ: $clique_3(10, 0.1)$\}, and the shaded areas visualize the standard deviations of each model's validation scores. From the figure, we noticed that for the latter 100 epochs, \textit{RL-KC} and \textit{RL-KV}'s validation performance oscillate significantly. Investigating the characteristics of \textit{Reinforcement Learning}, we discovered that RLSAT, upon encountering graphs with new scales, performs a whole new process of exploration. Therefore, RLSAT fails to generalize its learnt experience to subsequent larger graphs, which results in an unstable validation score during training. 
In contrast to RL-based model, our model adopts a highly parallel message-passing mechanism, which updates all nodes of all graphs simultaneously at each epoch.
% As a result, we are able to reach a high score rapidly, while maintaining the accuracy with little fluctuation.

In addition to testing on a diversified distribution of graph problems, we also experimented on the classic random 3-SAT dataset, and compared our results with that of PDP, which is recent work following NeuroSAT with a hybrid of GNN and RNN. As seen in Table \ref{tab:result_uf}, our model retains the ability to achieve a high clause assignment completion rate. In comparison, PDP takes a significantly longer time for inference, while reaching an average completion rate that does not exceed ours. To further analyze these speed discrepancies, we present in Figure \ref{fig:bar_chart} the average speedup of our model against that of PDP, with the performance of WalkSAT as the metric. As demonstrated in the figure, our model is capable of achieving higher average test speeds regardless of the graph structure. This
observation can be explained by the fact that our model allows communication within homogeneous
nodes, which provide all nodes with abundant semantic information when updating their states.
Therefore, our model requires fewer iterations of message passing, and achieves greater efficiency.

\begin{figure}
    \centering
    \includegraphics[width=0.48\textwidth]{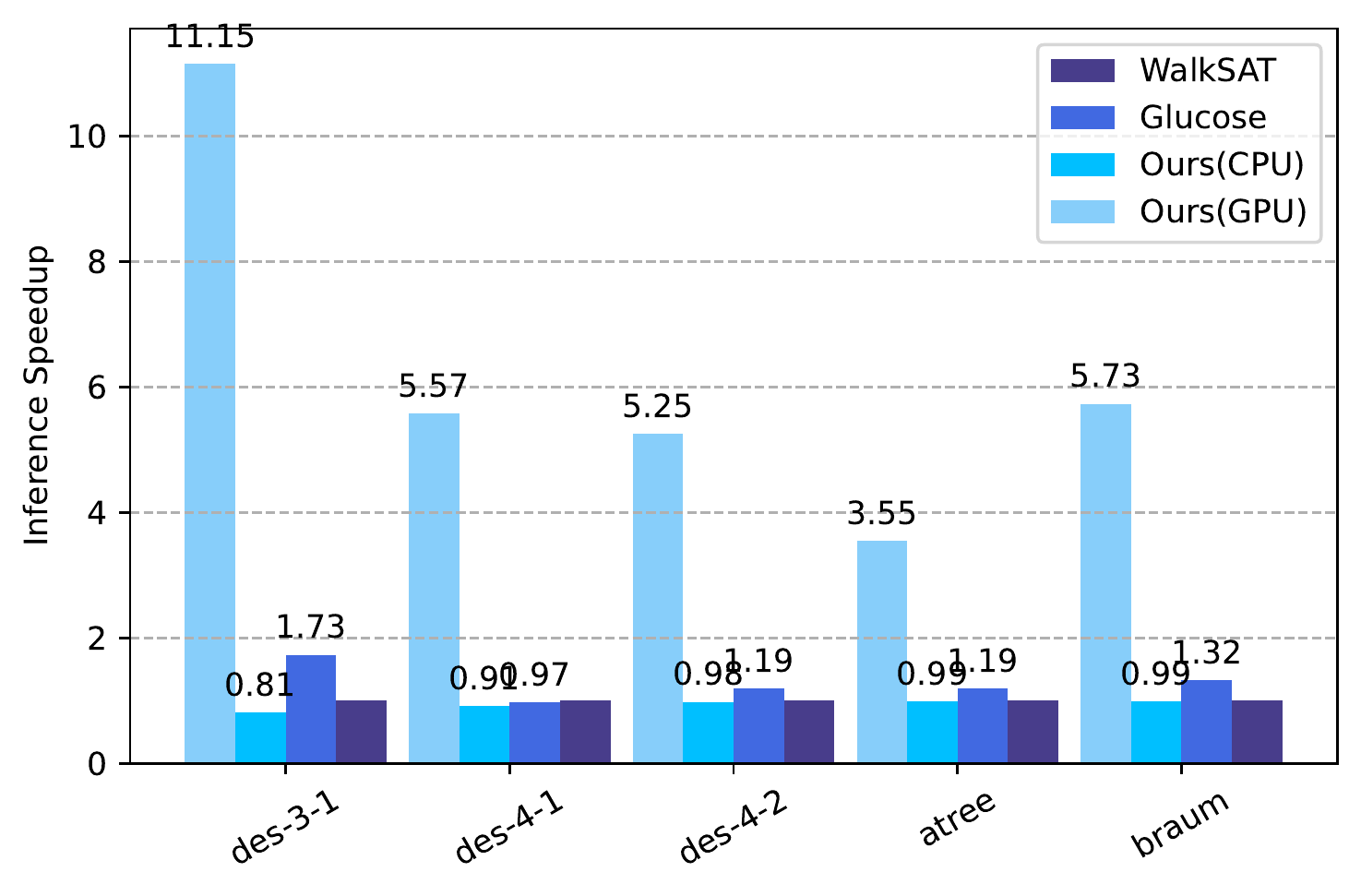} 
    \caption{
    The inference speedup for CNF-based logic synthesis.}
    \label{fig:bar_chart2}
\end{figure}

\subsubsection{\textbf{CNF-based logic synthesis}}
Apart from the general CSP/SAT evaluations, we also assessed our model's performance on solving CNF-based logic synthesis problems. Our proposed model is highly parallel and one-shot model based on neural symbolic learning for solving the CNF-based logic synthesis problems. Hence, we selected the classic stochastic algorithms WalkSAT and Glucose as authoritative baselines for comparison. We summarized the test results in Table \ref{tab:circuit_sat}. After training our model on the selected dataset for 500 epochs, our model achieves an average completion rate up to 88.7\% for circuit of DES datasets, and 89.3\% for arithmetic circuit datasets. We did not compare the completion rate of our model to those of the heuristic solvers, since they eventually solve all the problems without time limitation. Rather we focused on our model's latency to pursue acceleration, which could potentially help discovering early partial assignments to the heuristic solvers.

Consequently, our solver paired with a more guaranteed but slower deterministic solver, provides substantial overall speedup, while ensuring a solution. Visualized in Figure \ref{fig:bar_chart2} are our model's average solve speeds compared against that of Glucose, with the performance of WalkSAT as the metric. Once again, our model significantly outperforms the baseline models, regardless of the circuit structures being analyzed.
Furthermore, it is worth noting that our test set contains instances with very different distributions regarding variable and clause numbers, which reflects our model's scalability to work on wide range of tasks (as shown in Table\ref{tab:circuit_sat}) for logic synthesis problems of diverse difficulties without changing the main architecture.

% ------------------------------- Conclusion ---------------------------------
\section{Conclusion}
In this paper, we proposed \textit{Transformer-based SAT Solver (TRSAT)}, a one-shot model derived from the eminent Transformer architecture for bipartite graph structures, to solve the MaxSAT problem. We then extended this framework for logic synthesis task. We defined the homogeneous attention mechanism based on meta-paths for the self-attention between literals or clauses, as well as the heterogeneous cross-attention based on the bipartite graph links from literals to clauses, vice versa. Our model achieved exceptional parallelism and completion rate on the bipartite graph of MaxSAT with arbitrary sizes. The experimental results have demonstrated the competitive performance and generality of \textit{TRSAT} in several aspects. For future work, we want to analyze our initial results to check how the predicted partial solutions could contribute to the heuristic solvers to reduce the number of backtracking iteration in finding exact SAT solutions.

% ------------------------------- References ---------------------------------

\bibliographystyle{unsrt}
\bibliography{references.bib}

\end{document}